\documentclass{article} 
\usepackage{iclr2016_conference,times}
\usepackage{hyperref}
\usepackage{url}
\usepackage{amssymb}
\usepackage{amsmath}
\usepackage{multicol}
\usepackage{graphicx}
\title{Analyzing Stability of Convolutional Neural Networks in the Frequency Domain}

\author{Elnaz J. Heravi, Hamed H. Aghdam \& Domenec Puig\\
Computer Engineering and Mathematics Department\\
Rovira i Virgili University, 
Tarragona, 43007, Spain \\
\texttt{\{elnaz.jahani,hamed.habibi,domenec.puig\}@urv.cat} \\
}
\DeclareMathOperator*{\argmin}{arg\,min}
\DeclareMathOperator*{\argmax}{arg\,max}

\begin{document}

\maketitle

\begin{abstract}
Understanding the internal process of ConvNets is commonly done using visualization techniques. However, these techniques do not usually provide a tool for estimating the stability of a ConvNet against noise. In this paper, we show how to analyze a ConvNet in the frequency domain using a 4-dimensional visualization technique. Using the frequency domain analysis, we show the reason that a ConvNet might be sensitive to a very low magnitude additive noise. Our experiments on a few ConvNets trained on different datasets revealed that convolution kernels of a trained ConvNet usually pass most of the frequencies and they are not able to effectively eliminate the effect of high frequencies. Our next experiments shows that a convolution kernel which has a more concentrated frequency response could be more stable. Finally, we show that fine-tuning a ConvNet using a training set augmented with noisy images can produce more stable ConvNets.
\end{abstract}

\section{Introduction}
\label{sec:intro}

In the task of object recognition, the input of a Convolutional Neural Networks (ConvNets) is usually a $3$-channel image. Consequently, dimensions of the filters in the first convolution layer could be $w_1\times h_1 \times 3$. Assuming that the first layer consists of $K$ filters, the input to the second convolution layer might be a $K$-channel image where each channel is called a \textit{feature map}. Therefore, the dimensions of the filters might be $w_2\times h_2 \times K$. Since convolution filters are the main building block of ConvNets it is crucial to understand what happens when the input image is convolved using these filters. Also,  we may be able to decipher the function of each layer in a ConvNet by analyzing each filter separately. However, interpreting 3D filters is not trivial in the spatial domain. Specially, in the case of ConvNets, the third dimension of the filters is usually high since it depends on the number of the input channels which makes them harder to interpret.

There is a large body of work on understanding the internal process of ConvNets through visualization of hidden units. \citet{Zeiler2013} visualize the hidden units using Deconvolutional Networks. To be more specific, they reconstruct the images which have highly activated each unit. By this way, we can assess how each unit see the world and which parts of objects activate each neuron more. \citet{Simonyan2013} find a $L_2$-regularized image for each class by maximizing the class specific score. They also compute a class saliency map for the input image. 

\citet{Girshick2014} keep record of activations for a specific unit by entering many images to ConvNet and calculating their activations on the unit. Then, the images are sorted according to their activation on this particular unit and illustrated. Taking into account the fact that each unit in top layers has a corresponding receptive field on the image, it is possible to see which parts are important for each unit.

\citet{Mahendran2014} invert the \textit{d}-dimensional representation of an image computed by function $\Theta:\mathbb{R}^{H\times W\times C}\longrightarrow\mathbb{R}^d$. This approach tells us that to which extend it is possible to reconstruct the image using the representation function $\Theta$. By applying this method on each layer of the network we can understand which information is preserved by each layer. Similarly, \citet{Dosovitskiy2015} reconstructed the image by minimizing the squared Euclidean distance between the downsampled input image and reconstructed image. Recently, \citet{Nguyen2015} developed an evolutionary algorithm for generating images that do not look like to any of objects in the database but are classified with high score by ConvNet into one of object classes. 

Even though the visualization approaches help us to better understand the internal process of ConvNets, they do not provide a tool for assessing the stability of a ConvNet against noise. To address this problem, \citet{Szegedy2013} proposed a method for finding a $L_2$ regularized additive noise which minimizes the score of a specific class. 


\textbf{Contribution:} In practice, it is necessary to examine how stable are ConvNets when the input image is noisy. This is empirically achievable by evaluating a ConvNet using a contaminated test set. Another way is to analyze the filters in each layer in domains rather than the spatial domain. In this paper, we show how to analyze the filters of different layers in the frequency domain (Section \ref{sec:freq_domain}). To our knowledge, this is the first time that ConvNets are analyzed in the frequency domain. Then, we empirically assess various ConvNet architectures on different object recognition datasets (Section \ref{sec:experiments}). The experiments try to compare various choices for the \emph{loss} function, \emph{activations} and the \emph{input size}. Moreover, they illustrate that training a ConvNet using a noisy training set may increase the stability of the network. Above all, we analyze the ConvNets in the frequency domain to find out why all ConvNets are sensitive to small changes in the input.

\section{Analysis in the Frequency Domain}
\label{sec:freq_domain}
The response of convolution filters to an arbitrary noisy input can be analyzed \emph{empirically} or in the \emph{frequency} domain. For instance, consider two well-known $3\times3$ edge detection filters namely \emph{Sobel} and \emph{Prewitt}. In real applications, we are interested in an edge detection filter which is invariant to noise. To find out which of the above filters possess this property, we can generate hundreds of noisy images and apply each of these filters on the noisy inputs. Then, it is possible to compare the results with the ground truth images to quantitatively evaluate both filters. Although this is a valid approach, it has two important problems. 

First, it might be practically infeasible to use this method for analyzing the ConvNets trained on large-scale datasets. For example, it is not practical to use this method for evaluating the stability of the Googlenet~\citep{Szegedy2014} on the ImageNet dataset. This is due to the fact that the ImageNet dataset consists of 150K images. To empirically quantify the stability of the Googlenet, we must generate tens of noisy images with various signal-to-noise ratios (SNRs). If the images are degraded using a Gaussian noise with 10 different $\sigma$ values and 100 noisy images are generated for each value of $\sigma$, we will end up with $10\times100\times150K$ noisy images to be fetched into the Googlenet. This is not practically possible since it will take a very long time to complete this experiment. 

Second, the aforementioned empirical method does not tell us why a particular ConvNet is not invariant to noise. In other words, in order to make a ConvNet more tolerant to noise, we must know the reasons that make the ConvNet sensitive to image degradation.

More general approach is to transform convolution filters into the frequency domain and analyze their frequency response. High frequencies usually belong to noisy pixels. Hence, a convolution filter can be more invariant to noise if it does not respond to high frequency pixels. In the next section we review the Fourier transform for computing the frequency response of convolution filters.

\subsection{Review of The Fourier Transform}
\label{sec:fourier}
The Fourier transform decomposes a N-dimensional signal into N-dimensional \emph{sin} and \emph{cos} functions with various frequencies. The strength of each frequency is indicated by the magnitude of the \emph{sin} and \emph{cos} functions for that particular frequency. Mathematically, the Fourier transform of a 3-dimensional signal is defined as follows:
\begin{equation}
	\mathcal{F}(\mathcal{E}_1,\mathcal{E}_2,\mathcal{E}_3) = \int_{-\infty}^{\infty}\int_{-\infty}^{\infty}\int_{-\infty}^{\infty}e^{-2\pi i(x_1\mathcal{E}_1 + x_2\mathcal{E}_2 + x_3\mathcal{E}_3)}f(x_1, x_2, x_3) dx_1 dx_2 dx_3.
	\label{eq:fourier}
\end{equation}
In this equation, $\mathcal{E}_i$ is the frequency along $i^{th}$ axis and $f$ is a 3D signal. In the case of ConvNets, $f$ could be a 3D convolution kernel or a 3D feature map. $\mathcal{F}(\mathcal{E}_1,\mathcal{E}_2,\mathcal{E}_3)$ is a complex number indicating the magnitude and the phase of frequency triple $(\mathcal{E}_1,\mathcal{E}_2,\mathcal{E}_3)$ in the signal $f$.

\begin{figure}
	\centering
	\includegraphics[width=0.3\linewidth]{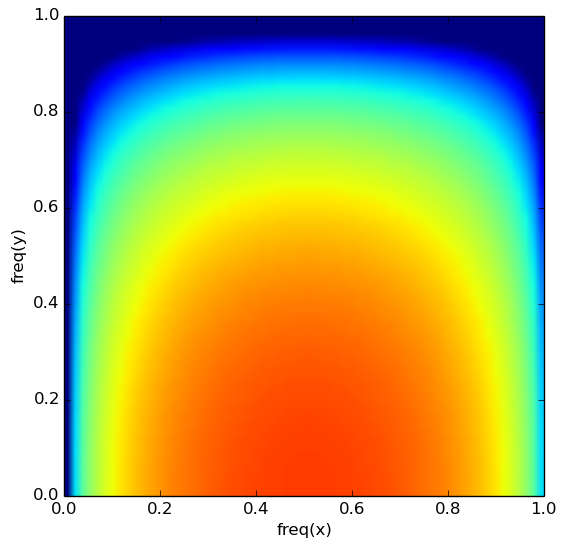}
	\includegraphics[width=0.3\linewidth]{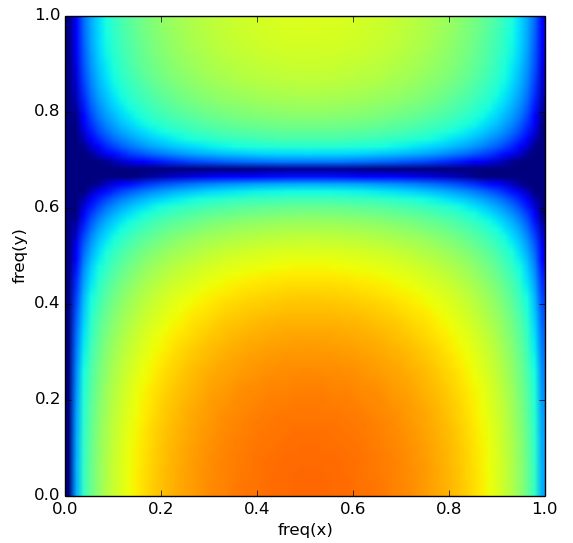}
	\caption{Frequency response of the Sobel (left) and the Prewitt (right) filters. The colder the color, the lower the magnitude (Best viewed in color).}
	\label{fig:sobel_prewitt}
\end{figure}

The frequency response of a filter/feature map can be obtained by computing (\ref{eq:fourier}) on every spatial location of the filter/feature map. Taking into account the fact that convolution in the spatial domain is equivalent to multiplication in the frequency domain, we realize that a convolution kernel could be more tolerant to noise if the magnitude of its frequency response is low for high frequencies. The Sobel filter is usually the best choice for calculating the first derivative of an image compared with other well-known $3\times3$ edge detection filters. To see the reason,  we reduced (\ref{eq:fourier}) into two dimensions and calculated the frequency response of the Sobel and the Prewitt filters\footnote{Filters are padded with zero to obtain a high resolution frequency response}. Figure \ref{fig:sobel_prewitt} illustrates the responses. 

It is clear that the Sobel filter (in X direction) decreases the effect of high frequencies along $y$ axis. In contrast, the Prewitt filter is not able to suppress high frequencies along $y$ axis. Taking into account that high frequencies are usually the result of noisy pixels, it shows that the Sobel filter is more tolerant against noise. For this reason, it is commonly the best $3\times3$ edge detection filter. 

\subsection{Frequency Response of ConvNets}
\label{sec:cnn_fft}
The filters of a ConvNet can be studied in the same way that we analyzed the Sobel and the Prewitt filters. The only difference is that filters of a ConvNet are usually 3D arrays so they must be visualized using 4D visualization techniques. \citet{Szegedy2013} showed that adding a low magnitude noise to an image which is barely perceivable to human eye may cause a ConvNet to incorrectly classify the noisy image. We can look for the reason in the frequency domain. To this end, it is enough to only study the effect of the additive noise. This is due to the linearity property of the Fourier transform. In other words, representing the image and the noise by $f$ and $r$, respectively, linearity property shows that the Fourier transform of the noisy image can be found by separately calculating the Fourier transform of image $f$ and noise $r$ and adding their results. Mathematically:
\begin{equation}
	\mathcal{F} (\alpha f + \beta r) = \alpha\mathcal{F}(f) + \beta\mathcal{F}(r).
	\label{eq:fourier_linearity}
\end{equation}
Therefore, we only need to transform the noise into the frequency domain in order to analyze the effect of the additive noise on the output of a ConvNet. This is derived by the fact that $\mathcal{F} (f + r)-\mathcal{F} (f) = \mathcal{F} (r)$. 

\begin{figure}
	\centering
	\includegraphics[width=0.95\linewidth]{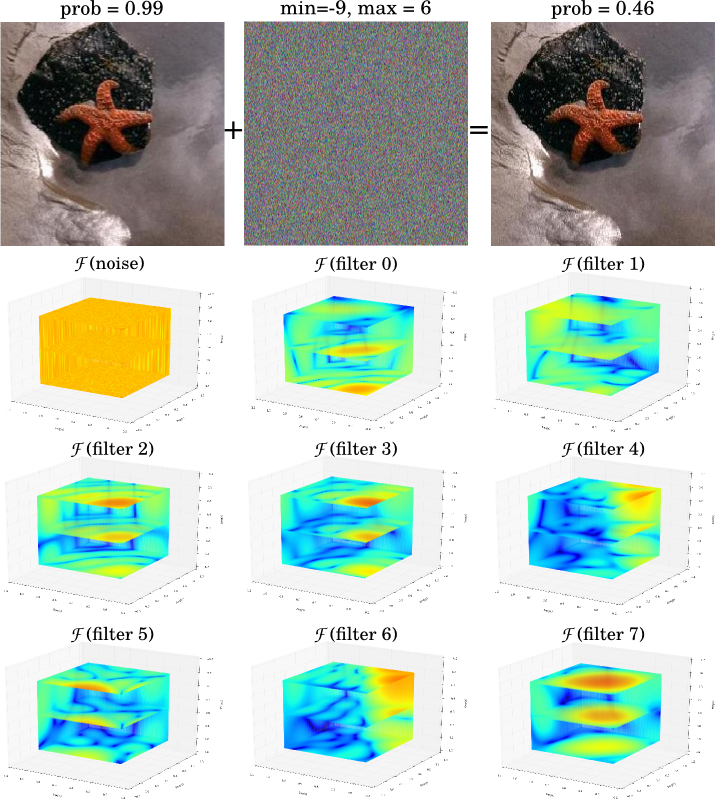}
	\caption{Analyzing the minimum noise in the frequency domain using the first 7 filters in the first layer of the Googlenet 
	obtained from \citet{Jia2014}. The intensity of noise has been normalized so it is perceivable to human eye. The colder the color, the smaller the spectrum (Best viewed in color). }
	\label{fig:gaussian_noise}
\end{figure}

Our goal is to find out why a low magnitude noise may cause a ConvNet to incorrectly classify an image. For this purpose, we consider the pre-trained model of the Googlenet~\citep{Szegedy2014} provided by \citet{Jia2014}. Then, it is fine-tuned on the Caltech101~\citep{Fergus2004} dataset by adjusting the weights in the classification layer and freezing the weights in the other layers. Finally, an additive noise is found by minimizing the following objective function:
\begin{equation}
	r^* = \argmin_r\;\; \psi(loss(\mathcal{X}+r), c, k) + \lambda\|r\|_2\\
	\label{eq:cnn_stability_our_1}
\end{equation}
\begin{equation}
	\psi(\mathcal{L}, c, k)=
	\left\{
	\begin{matrix}
		\beta\times\mathcal{L}[c] & \argmax \mathcal{L} = c \\
		\mathcal{L}[k]-\mathcal{L}[c] & otherwise
	\end{matrix}\right.
	\label{eq:cnn_stability_our_2}
\end{equation}
\noindent where $c$ is the actual class label, $k$ is the predicted class label, $\lambda$ is the regularizing weight and $loss(\mathcal{X}+r)$ returns the loss vector of the degraded image $\mathcal{X}+r$ computed over all classes. Also, $\beta$ is a multiplier to penalize those values of $r$ that do not properly degrade the image so it is not misclassified by ConvNet. We minimized the above objective function on a sample image from the Calteach101 dataset. Figure \ref{fig:gaussian_noise} illustrates the frequency response of $r$ along with the frequency response of the first 7 filters in the first layer of the Googlenet. Note that the maximum and minimum values of the noise are very small. However, their intensity are normalized for visualization purposes.

First, we observe that the noise affects almost all the frequencies (Note that on the chart, only points with blue color shows a magnitude near zero).  Second, the frequency responses of the filters reveal that not only they pass low and mid frequencies they also pass high frequencies. If the frequency response of each filter is multiplied with the response of the noise (\emph{i.e.} convolution in spatial domain), the result will be another noisy image where the effect of some frequencies are slightly reduced. In other words, the output of the first convolution layer in the Googlenet is a multi-channel noisy image since the filters are not able to suppress the effect of the noise. 

When the noisy multi-channel image is passed through a max-pooling layer, it may produce another noisy image where the magnitude of high frequencies may increase. 
Analyzing several ConvNets in frequency domain shows that they tend to learn filters which respond to most of the frequencies in the image. For this reason, the noise is propagated along the network and they also appear in the last convolution layer where they may alter the output of the ConvNet. 

Now the question is can we solve this problem? From the frequency domain perspective, it is not trivial to suppress the additive noise $r$ during the convolution process. Because $r$ has positive magnitude in nearly all the frequencies. Hence, even discarding effect of the noise on some frequencies is not going to effectively solve the problem since the frequencies which correspond to noise will be passed to the next layers through other frequencies. However, as we will show in the next section, by learning filters which are more localized in the frequency domain, the stability of the network may increase while the accuracy of the network remains the same.

\section{Experiments}
\label{sec:experiments}
In this section, we study the stability of ConvNets empirically and in the frequency domain. To this end, we utilize ConvNets with different architectures trained on various datasets. Specifically, we use the architecture in \citet{Jia2014} for training a ConvNet on the CIFAR10 dataset ~\citep{Krizhevsky2009}. We also use the pre-trained models of Alexnet~\citep{Krizhevsky2012} and Googlenet~\citep{Szegedy2014} from \citet{Jia2014} and fine-tune them on the Caltech101 dataset~\citep{Fergus2004}. Finally, we train the architectures from \citet{Ciresan2012a} and \citet{Aghdam2015} on the GTSRB~\citep{Stallkamp2012} dataset. Table \ref{tbl:overall_accuracy} shows the accuracy of each ConvNets trained on the original datasets. It is clear that all the ConvNets have achieved state-of-art results. 

\begin{table}
	\centering
	\small
	\caption{Accuracy of the benchmark ConvNets on the original datasets. Trained models of the AlexNet
	and the Googlenet
	as well as the architecture of the CIFAR10 ConvNet have been obtained from \citet{Jia2014}. The architecture of the IDSIA and the IRCV ConvNets have been obtained from \citet{Ciresan2012a} and \citet{Aghdam2015}, respectively.}
	\label{tbl:overall_accuracy}
	\begin{tabular}{|l|c|l|c|}
		\hline
		Network 				& accuracy (\%) & Network 					& accuracy (\%)\\
		\hline
		CIFAR10 (hing+relu)		 	& 79.8 			& CIFAR10 (soft+relu) 		& 78.6\\		
		IRCV (GTSRB)(soft+relu)			& 99.01 		& IDSIA (GTSRB)(soft+tanh) 				& 98.77 \\		
		Alexnet (Caltech101)(soft+relu)	& 87.39			& Googlenet (Caltech101)(soft+relu)	& 91.51	\\		
    	 \hline
	\end{tabular}	
\end{table}


\subsection{Stability of ConvNets}
\label{sec:exp_stability}
To empirically study the stability of the ConvNets against noise, the following procedure is conducted. First, we pick the test images from the original datasets which are \textit{correctly classified} by the ConvNets. Then, $100$ noisy images are generated for each $\sigma \in \{1, 2, 4, 8, 10, 15, 20, 25, 30, 35, 40\}$. In other words, $1100$ noisy images are generated for each of correctly classified test images from the original datasets. The same procedure is repeated on every dataset and the accuracy of the ConvNets is computed using the noisy test sets. Table \ref{tbl:stability_noaugment} shows the accuracy of the ConvNets per each value of $\sigma$.

\begin{table}
	\centering
	\small
	\caption{Accuracy of the ConvNets obtained by degrading the \textit{correctly classified test images} in the original datasets using a Gaussian noise with various values of $\sigma$.}
	\label{tbl:stability_noaugment}
	\begin{tabular}{|l|c|c|c|c|c|c|c|c|c|c|c|}
		\cline{2-12}
		\multicolumn{1}{c}{} & \multicolumn{11}{|c|}{accuracy (\%) for different values of $\sigma$}  \\
		 \hline
		 network & 1 & 2 & 4 & 8 & 10 & 15 & 20 & 25 & 30 & 35 & 40 \\
		 \hline
		   IRCV  & 100.0 &  100.0 &  99.8 &  99.3 &  98.8 &  97.4 &  94.3 &  91.2 &  87.8 &  84.5 &  81.4 \\		   		    
		   IDSIA  & 99.9 &  99.9 &  99.7 &  99.0 &  98.5 &  97.1 &  94.2 &  91.2 &  88.0 &  84.7 &  81.6  \\
		 CIFAR10 (hing)& 99.7 &  99.2 &  98.0 &  94.4 &  91.7 &  84.7 &  71.7 &  59.5 &  47.6 &  37.7 &  30.1  \\		
		 CIFAR10 (soft)& 99.7 &  99.3 &  98.3 &  95.4 &  93.6 &  88.4 &  77.7 &  67.8 &  58.2 &  49.7 &  42.4  \\
 		 Alexnet & 100.0 &  99.9 &  99.6 &  98.7 &  97.7 &  95.7 &  91.4 &  86.7 &  80.5 &  73.0 &  65.2   \\
   	   Googlenet & 99.8 &  99.7 &  99.5 &  98.5 &  97.8 &  96.0 &  92.7 &  89.2 &  85.1 &  80.3 &  75.2   \\	
		 \hline
	\end{tabular}	
\end{table}

First, we observe that except the IRCV and the Alexnet other ConvNets have misclassified a few of the \textit{correctly classified} test images which are degraded using a Gaussian noise with $\sigma=1$. Note that when $\sigma=1$, it is highly improbable that a pixel is degraded more than $\pm4$  intensity levels in each channel. However, this slight change in the input can lead some of the ConvNets to incorrectly classify the image. Also, as the value of $\sigma$ increases, the accuracy of the ConvNets reduces. This is consistent with the explanation in Section \ref{sec:cnn_fft} in the sense that a higher value of $\sigma$ increases the magnitude of the all frequencies. Since the convolution layers are not able to effectively reduce the noise, they are propagated through the ConvNet and alter the output of final convolution layer.

Second, a squashing activation function such as \textit{tanh} seems to be more tolerant against noise since it maps the input with higher values to the outputs with very close values. However, comparing the results obtained from the IRCV (relu) and the IDSIA (tanh) illustrate that a squashing activation function does not necessarily make a ConvNet more robust against additive noise. 

Third, comparing the results from the CIFAR10 ConvNets trained using the \textit{softmax} and the \textit{hing} loss functions illustrate there is not a golden rule that a specific loss function leads to a more stable ConvNet. We observe that both ConvNets makes mistakes even when $\sigma=1$.

Fourth, it is observable that there is not a clear relation between the size of the input and the stability of the ConvNet. To be more specific, the size of the input to the IDSIA and the IRCV ConvNets is $48\times48$ pixels and it is $32\times32$ pixels in the case of the CIFAR10 ConvNets. Moreover, the size of the inputs of the Alexnet and the Googlenet is $227\times227$ and $224\times224$ pixels, respectively. Notwithstanding, the IRCV and the IDSIA are more stable than the Alexnet and the Googlenet. This is due to the fact that objects in the GTSRB dataset are simpler than the objects in the ImageNet dataset. In addition, the number of classes in the GTSRB dataset is much less than the number of the classes in the ImageNet dataset. For these reasons, a $48\times48$ is enough for the IRCV and the IDSIA ConvNets to learn reasonably stable ConvNets. In contrast, the CIFAR10 dataset contains complex objects which are presented in small images. For this reason, some important details of the objects are missed due to down-sampling. When the images are degraded by a strong noise, it dramatically changes the frequency pattern which in turn alters the classification score. In sum, stability of a ConvNet does not solely depend on the size of the input. Instead, choosing an appropriate input size according to the number of the classes and complexity of the objects in the dataset can increase the stability of a ConvNet.

\subsection{Smoothing the noisy images}
\label{sec:smoothing}
It is a common practice to smooth an image using a Gaussian filter before further processing. In this experiment, we investigate how smoothing affects the stability of the ConvNets. For this purpose, we follow the same procedure as in Section \ref{sec:exp_stability} but the noisy images are smoothed using a $3\times3$ Gaussian filter with $\sigma=0.5$ before feeding to the ConvNets. Table \ref{tbl:smooth} shows the results.

\begin{table}
	\centering
	\small
	\caption{Accuracy of the ConvNets obtained by smoothing the noisy inputs using a $3\times3$ Gaussian filter with $\sigma=0.5$.}
	\label{tbl:smooth}
	\begin{tabular}{|l|c|c|c|c|c|c|c|c|c|c|c|}
		\cline{2-12}
		\multicolumn{1}{c}{} & \multicolumn{11}{|c|}{accuracy (\%) for different values of $\sigma$}  \\
		 \hline
		 network & 1 & 2 & 4 & 8 & 10 & 15 & 20 & 25 & 30 & 35 & 40 \\
		 \hline
		   GTSRB & 99.9 &  99.9 &  99.8 &  99.3 &  98.9 &  97.7 &  95.2 &  92.6 &  89.6 &  86.5 &  83.6  \\		   		    		  
		 CIFAR10 (soft)& 90.7 &  90.6 &  90.6 &  90.4 &  90.3 &  89.5 &  86.8 &  82.9 &  77.7 &  71.7 &  65.3  \\
 		 Alexnet & 98.4 &  98.4 &  98.5 &  98.2 &  97.8 &  96.5 &  93.7 &  90.4 &  85.6 &  79.9 &  73.8  \\
		 \hline
	\end{tabular}	
\end{table}

\begin{figure}
	\centering
	\includegraphics[scale=0.20]{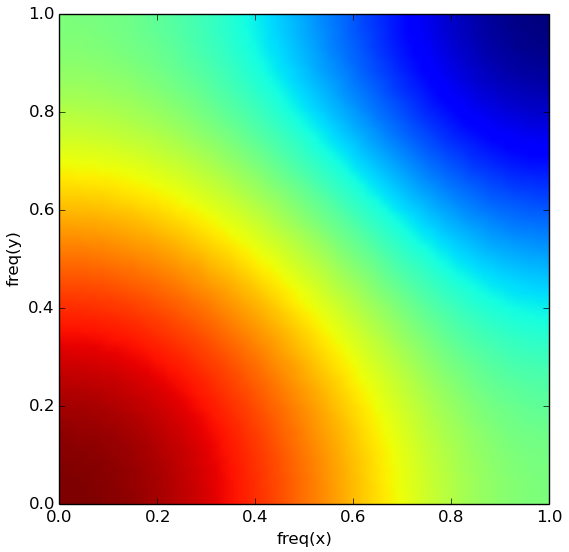}
	\caption{Frequency response of a $3\times3$ Gaussian filter with $\sigma=0.5$. The colder the color, the smaller the magnitude (Best viewed in color).}
	\label{fig:gaussian_filter_fft}
\end{figure}

In general, smoothing images that are degraded using the Gaussian noise with $\sigma \le 10$ has a negative impact on the accuracy. In particular, it dramatically reduces the performance of the CIFAR10 ConvNets (when $\sigma \le 10$). Because, images in the CIFAR10 dataset are very small and smoothing a clean image eliminates some of the important details of the objects. In contrast, smoothing images degraded by a Gaussian noise with $\sigma \ge 15$ increases the overall accuracy of the ConvNets. The reason for this behaviour can be found in the frequency domain. Figure \ref{fig:gaussian_filter_fft} illustrates the frequency response of the Gaussian filter.

According to this figure, the Gaussian filter passes the low frequencies but if slightly reduces the effect of mid frequencies. In addition, it significantly reduces the effect of high frequencies. When, the image is degraded by a Gaussian noise with small $\sigma$ values, it might not significantly affect the mid and the low frequencies. However, applying a Gaussian filter on these images affect their mid frequencies considerably which causes that some of the images are incorrectly classified. In other words, the effect of Gaussian smoothing on the mid frequencies is more than the effect of noise on the same frequencies. In contrary, mid and high frequencies are dramatically affected in images degraded with higher values of $\sigma$. As the result, applying a Gaussian smoothing filter may considerably reduce their effect which in sequel increases the accuracy of the ConvNet on these images. The immediate conclusion from this experiment is that if the filters of the ConvNets are further adjusted to weaken high frequencies and to compress the frequency response, the stability of the ConvNets may increase when the input images are noisy.

\subsection{Augmenting with noisy images}
\label{sec:augment}
Augmenting data by applying some transformations on the original dataset is a common practice for increasing the generalization of ConvNets. The data augmentation procedure does not usually involve adding noisy images to a dataset. In this experiment, we augment the original dataset with noisy images which are generated using the Gaussian noise. We consider $\sigma \in \{1, 5, 10, 20\}$ and 10 different noisy images are generated for each sample in the original training set. Next, \textit{all fully connected layers after the last convolution layer are frozen and only convolution layers are trained using the noisy datasets}. By this way, fine-tuning procedure only affects the convolution filters. Finally, the ConvNets are evaluated by creating a noisy test set as we mentioned in Section \ref{sec:exp_stability}. Table \ref{tbl:accuracy_noisy} and Table \ref{tbl:stability_augment} show the accuracy of the ConvNets obtained by applying on the original test set and the noisy test set, respectively. As it is clear from Table \ref{tbl:accuracy_noisy}, the ConvNets have achieved very close accuracies compared with Table \ref{tbl:overall_accuracy}.
 
 \begin{table}
 	\centering
 	\small
 	\caption{Accuracy of the ConvNets trained by the noisy datasets and tested by the original test set.}
 	\label{tbl:accuracy_noisy}
 	\begin{tabular}{|l|c|l|c|}
 		\hline
 		Network 				& accuracy (\%) & Network 					& accuracy (\%)\\
 		\hline
 		IRCV (noisy) 			& 99.29  		& IDSIA (noisy) 			& 98.59\\
 		CIFAR10 (noisy+hing)	& 78.2 			& cifar10 (noisy+softmax) 	& 76.6\\ 		
     	 \hline
 	\end{tabular}	
 \end{table}
 
\begin{table}
	\centering
	\small
	\caption{Accuracy of the ConvNets after augmenting the original dataset with noisy images degraded by the Gaussian noise with $\sigma \in \{1, 5, 10, 20\}$.}
	\label{tbl:stability_augment}
	\begin{tabular}{|l|c|c|c|c|c|c|c|c|c|c|c|}
		\cline{2-12}
		\multicolumn{1}{c}{} & \multicolumn{11}{|c|}{accuracy (\%) for different values of $\sigma$}  \\
		 \hline
		 network & 1 & 2 & 4 & 8 & 10 & 15 & 20 & 25 & 30 & 35 & 40 \\
		 \hline
		   IRCV & 100.0 &  99.9 &  99.9 &  99.5 &  99.2 &  98.5 &  96.8 &  94.8 &  92.5 &  89.8 &  87.2  \\
		   IDSIA & 99.9 &  99.9 &  99.7 &  99.2 &  98.9 &  98.0 &  96.1 &  94.1 &  91.9 &  89.4 &  87.0  \\
 		   CIFAR10 (hing)& 99.8 &  99.6 &  99.3 &  98.3 &  97.6 &  96.3 &  94.2 &  92.2 &  89.9 &  87.7 &  85.0  \\ 
		   CIFAR10 (soft)& 99.6 &  99.4 &  98.8 &  97.6 &  96.9 &  95.5 &  93.2 &  91.1 &  88.7 &  86.4 &  83.7  \\
		 \hline
	\end{tabular}	
\end{table}

\begin{figure}[!tbh]
	\centering
	\includegraphics[width=0.31\linewidth]{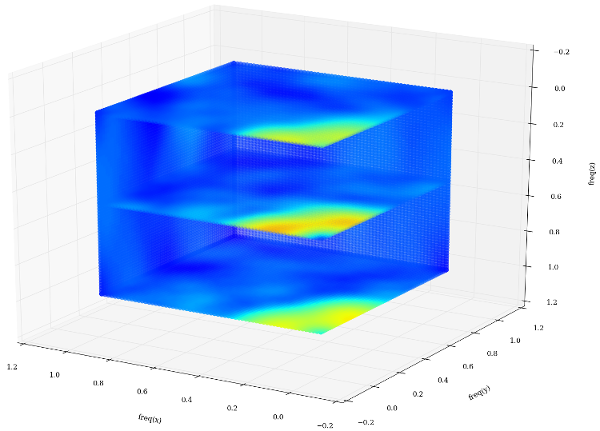}
	\includegraphics[width=0.31\linewidth]{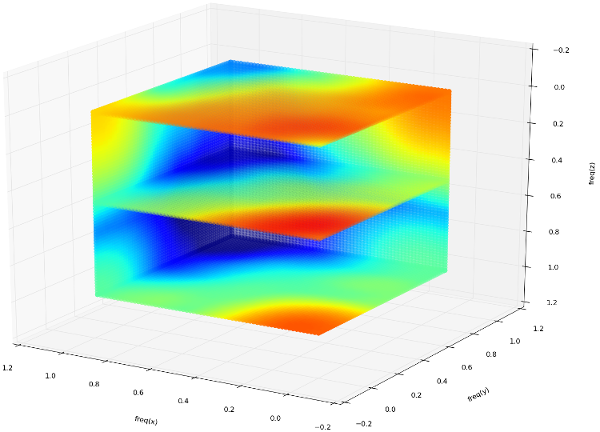}
	\includegraphics[width=0.31\linewidth]{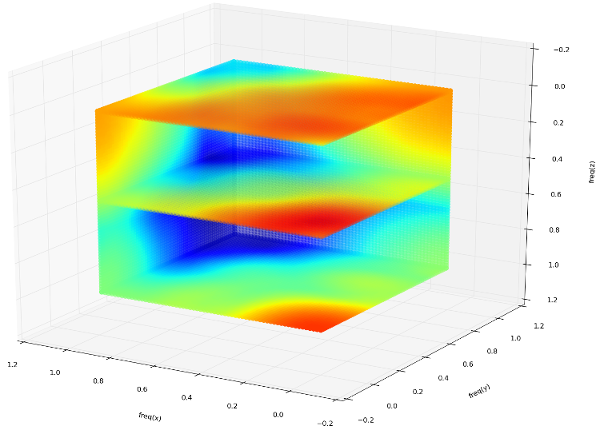}
	
	\includegraphics[width=0.31\linewidth]{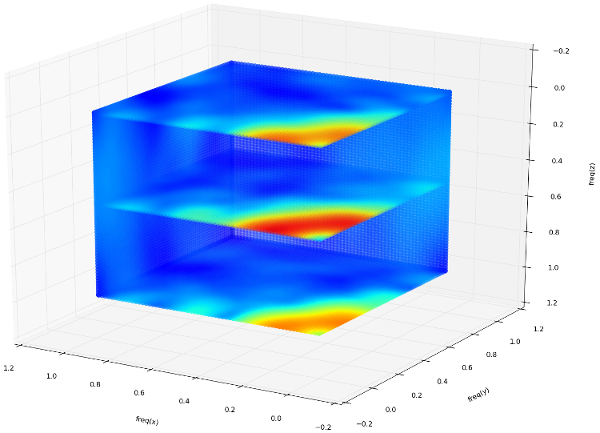}
	\includegraphics[width=0.31\linewidth]{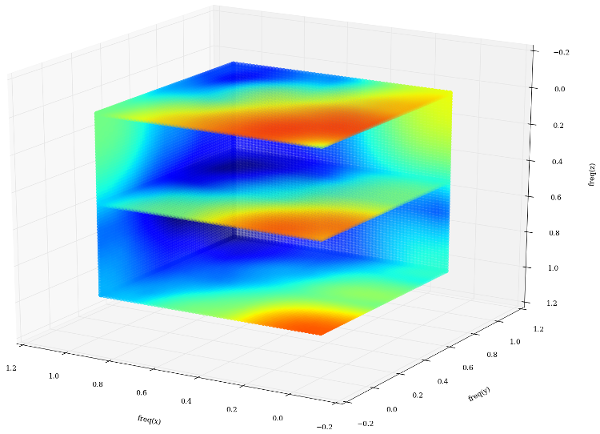}
	\includegraphics[width=0.31\linewidth]{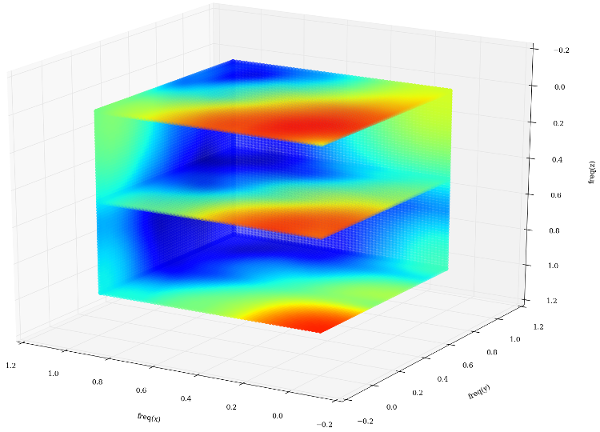}
	\caption{The mean spectrum of the first layer of the IRCV (left), the CIFAR10 (softmax, middle) and the CIFAR10 (hing, right) ConvNets trained using the original (top) and the noisy (bottom) training datasets (Best viewed in color).}
	\label{fig:mean_fft_cifar_soft}
\end{figure}

The results illustrate a considerable increase in the accuracy of the ConvNets, especially on the images degraded by a strong Gaussian noise. In contrast to the previous experiment (\emph{i.e.} smoothing using a Gaussian filter), the fine-tuned convolution kernels do not have a negative effect on the images degraded by a Gaussian noise with $\sigma \le 10$. However, it is clear that ConvNets learn to reduce the effect a strong noise. To investigate this issue, we computed the frequency response of the \textit{first layer} on the IRCV and the CIFAR10 (softmax and hing) ConvNets before and after augmenting the training set with noisy images. Then, the \textit{mean spectrum} of first layer of each ConvNet was computed, separately. Figure \ref{fig:mean_fft_cifar_soft} shows the results.
 
The common point in all plots is that the mean spectrum of the ConvNets fine-tuned using the noisy training set is more localized than the ConvNets trained without noisy images. In other words, a fewer frequencies are passed through the convolution filters trained using the noisy training set. For this reason, these ConvNets have the ability to effectively reduce the additive noise compared with the ConvNets that are trained using only the original dataset. In sum, augmenting a dataset using noisy images is advantageous and they help the training algorithm to learn the convolution filters with more concentrated spectrum. 

\section{Conclusion}
\label{sec:conclusion}
In this paper, we studied the stability of Convolutional Neural Networks (ConvNets) against image degradation. To this end, we showed how to analyze the convolution filters in a ConvNet by visualizing their Fourier transform in 4-dimensions. Then, we studied why a ConvNet may make mistakes by degrading the image using an additive noise which is barely perceivable to human eye. Specifically, we illustrated that an additive noise affects almost all the frequencies on the image. On the other hand, analyzing the convolution kernels in the frequency domain revealed that they may respond to high frequencies as mush as mid and low frequencies. For this reason, they are not able to effectively denoise the image and the noise is propagated across the ConvNet that alters the classification score. Moreover, our experiments on ConvNets trained on different datasets showed that there is not a golden rule to say a particular loss function or activation function yields a more stable ConvNet. Besides, the size of the input image can only affect the performance if it is not selected based on the complexity of the objects in the dataset and the number of the classes. Next, we studied what happens if the noisy image is smoothed using a Gaussian smoothing kernel. The results suggest that while it can be helpful on images with low signal-to-noise ratio but it can adversely affect the results on images with high signal-to-noise ratio. The frequency response of the Gaussian kernel show that it reduces the effect of mid frequencies. When it is applied on a clean image it manipulates the mid frequencies considerably which in turn changes the output. Similarly, when the image is degraded using a strong noise, the Gaussian kernel is able to effectively reduce the effect of the noise and more images are correctly classified. The outcome of this experiment is that if convolution kernels are trained properly to have a more concentrated frequency response it may increase the stability of the ConvNet. Finally, we investigated this assumption by augmenting the training set using noisy images. Applying the ConvNets trained using noisy sets on the noisy test sets illustrated a considerable performance boost. We analyzed the reason by computing the mean spectrum (frequency response) of the convolution filters in the first layer of the ConvNets before and after training by the noisy sets. It showed that the frequency response of the ConvNets training on noisy sets are more concentrated than the ConvNets trained on the clean set.

\subsubsection*{Acknowledgments}
Hamed H. Aghdam is grateful for the support granted by Generalitat de Catalunya's Ag\`{e}cia de Gesti\'{o} d'Ajuts Universitaris i de Recerca (AGAUR) through FI-DGR 2015 fellowship.

\bibliography{CNN-Analysis,CNN-Dataset,CNN-Library,CNN-TSR_IJCV,CNN-Visualization,CNN-My_papers}
\bibliographystyle{iclr2016_conference}

\end{document}